\documentclass{l4dc2025}

\title[Bridging Adaptivity and Safety]{Bridging Adaptivity and Safety: \\ Learning Agile Collision-Free Locomotion Across Varied Physics}

\author{... 
\thanks{*This work was not supported by any organization}
\thanks{$^{1}$For more info, please refer to the official project website {\tt\small\href{https://sites.google.com/andrew.cmu.edu/bas}{https://adaptive-safe-locomotion.github.io}}}%
}

\usepackage{float}
\usepackage{amsfonts, setspace, wrapfig,algorithm}
\usepackage{caption}
\usepackage{xcolor}
\usepackage{hyperref}
\usepackage{times}
\usepackage[skip=6pt plus1pt, indent=12pt]{parskip}
\usepackage{multicol}
\usepackage{multirow}
\usepackage{cleveref}
\usepackage[export]{adjustbox}

\author{%
 \Name{Yichao Zhong} \Email{yichaoz@andrew.cmu.edu}\\
 \addr Carnegie Mellon University
 \AND
 \Name{Chong Zhang} \Email{chozhang@ethz.ch}\\
 \addr ETH Zurich%
  \AND
 \Name{Tairan He} 
 \Email{tairanh@andrew.cmu.edu}\\
 \addr Carnegie Mellon University%
  \AND
 \Name{Guanya Shi} 
 \Email{guanyas@andrew.cmu.edu}\\
 \addr Carnegie Mellon University%
}

\begin{document}

\maketitle
\singlespacing
\vspace{-30pt}

\begin{abstract}

Real-world legged locomotion systems often need to reconcile agility and safety for different scenarios.
Moreover, the underlying dynamics are often unknown and time-variant (e.g., payload, friction). In this paper, we introduce BAS (\underline{B}ridging \underline{A}daptivity and \underline{S}afety), which builds upon the pipeline of prior work Agile But Safe (ABS)~\citep{he2024agilesafelearningcollisionfree} and is designed to provide adaptive safety even in dynamic environments with uncertainties. 
BAS involves an agile policy to avoid obstacles rapidly and a recovery policy to prevent collisions, a physical parameter estimator that is concurrently trained with agile policy, and a learned control-theoretic RA (reach-avoid) value network that governs the policy switch. Also, the agile policy and RA network are both conditioned on physical parameters to make them adaptive. 
To mitigate the distribution shift issue, we further introduce an on-policy fine-tuning phase for the estimator to enhance its robustness and accuracy.
The simulation results show that BAS achieves 50\% better safety than baselines in dynamic environments while maintaining a higher speed on average.
In real-world experiments, BAS shows its capability in complex environments with unknown physics (e.g., slippery floors with unknown frictions, unknown payloads up to 8kg), while baselines lack adaptivity, leading to collisions or degraded agility. As a result, BAS achieves a 19.8\% increase in speed and gets a 2.36 times lower collision rate than ABS in the real world.
Videos: \href{https://adaptive-safe-locomotion.github.io}{https://adaptive-safe-locomotion.github.io}.
\end{abstract}


\begin{figure}[h]
    \centering
    \includegraphics[width=\linewidth]{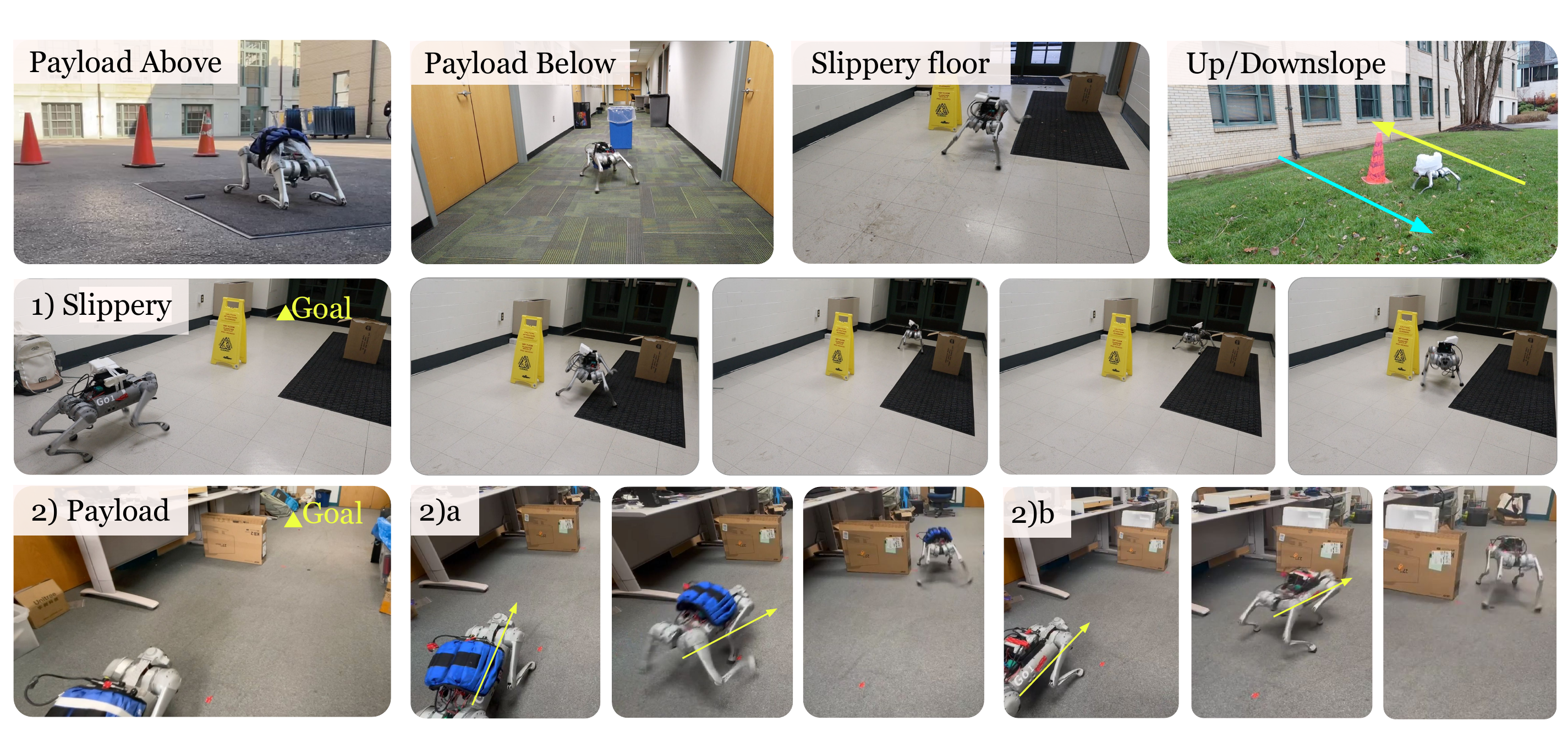}
    \caption{
    1) The robot can handle collision-free locomotion in even super slippery terrain condition (soap water on both floor and robot feet), and also can adapt to rough terrain (dry carpet) suddenly.  2) Adaptive recovery triggering of the robot in different circumstances, such as a) early recovery with 8kg payload and  b) late recovery with no payload. 
    }
    \label{fig:teaser}
\end{figure}

\section{INTRODUCTION}

Legged robot locomotion in cluttered and dynamic environments requires adaptivity to varying physics and environmental changes while simultaneously ensuring agility for efficient navigation and safety for reliable deployment. 
And such adaptivity to varied environments is claimed to be crucial for real-world tasks such as disaster response in forests~\citep{20203504172}, evacuation in fire-prone areas~\citep{10014987}, and rescue operations~\citep{9670087}.
Despite recent progress in legged locomotion~\citep{annurev:/content/journals/10.1146/annurev-control-042920-020211, Hwangbo2019, kumar2021rmarapidmotoradaptation, Lee2020, li2024reinforcementlearningversatiledynamic,xue2024full,he2024omnih2o,zhang2024wococo}, there remains a significant gap in methodologies that effectively integrate adaptivity, safety, and agility. 
In this work, we enable the robot to jointly achieve agility and safety with adaptivity, maintaining strong performance in challenging environments.

Striking a good balance of adaptivity, safety, and agility in legged locomotion remains a significant challenge, as focusing on one aspect often comes at the expense of the others.
Recent pioneer legged / wheeled locomotion works use reinforcement learning (RL)~\citep{levine2020offlinereinforcementlearningtutorial,silvermastering2017} to prioritize adaptability in agility to handle environmental changes~\citep{kumar2021rmarapidmotoradaptation,Lee2020,wang2024understandingkeyestimationlearning, long2023him, zhang2024catchitlearningcatch,yang2023cajuncontinuousadaptivejumping,luo2024pieparkourimplicitexplicitlearning,xiao2024anycar}. However, these approaches somehow neglect safety considerations.
On the other hand, the prior work ABS~\citep{he2024agilesafelearningcollisionfree} jointly pushes the safety and agility limits in nominal environments but is not adaptive to varying physics, and agility and safety performance can drop severely in challenging environments, as shown in our results in \Cref{fig:real_adapt}. 

Other studies~\citep{xiao2024safedeeppolicyadaptation,chiu2022collisionfreempcwholebodydynamic,yun2024safecontrolquadrupedvarying,Borquez2023,gao2024coohoi} focus on adaptivity and safety, but sacrifice agility or need precise dynamics in the real world. For example, \cite{yun2024safecontrolquadrupedvarying} solves safe-legged locomotion using reduced-order dynamics models, so it sacrifices much on speed. Moreover, on the theoretical side, \cite{Borquez2023} proves that it is possible to achieve adaptive safety by parameter-conditioned reachability analysis, but the ground truth physical parameters are not accessible in the real world.

In addition to being adaptive, another way to handle changing environments is to improve robustness by making the system more conservative~\citep{https://doi.org/10.1002/rob.21974,9196777}. However, being conservative can be insufficient in certain scenarios (e.g., search and rescue tasks).
Moreover, although the safety-related literature is rich~\citep{achiam2017constrainedpolicyoptimization,bansal2017hamiltonjacobireachabilitybriefoverview,liu2022constrainedvariationalpolicyoptimization,xu2021crponewapproachsafe,5685555,Hsu2023,liu2020robust}, most of them are not tested in the real world.
In summary, there is a missing space for adaptive, safe, and agile locomotion for the needs of real-world applications.

To address this, we propose BAS that builds on ABS (Agile But Safe)~\citep{he2024agilesafelearningcollisionfree} and manages to enhance the adaptivity to strike a balance.
Previously, ABS involves an agile policy to avoid obstacles rapidly and a recovery policy to prevent
failures, and a learned
control-theoretic reach-avoid value network, which governs the policy switch, guides the recovery policy as an objective function and safeguards the robot in a closed loop.
Yet, unlike ABS, BAS employs an explicit physics-parameter estimator learned from proprioceptive history during policy training as an adaptation module and feeds forward the estimated parameters to the controller and the RA (Reach-Avoid) network to enhance the adaptivity. To mitigate the distribution shift caused by switching between agile and recovery policies, we further introduce an end-to-end on-policy fine-tuning strategy, improving the accuracy of the estimator during inference. Extensive evaluations demonstrate that BAS significantly outperforms ABS and other adaptive-and-safe baselines in both safety and agility metrics. In real-world experiments, BAS achieves a 19.8\% advantage in speed and is 2.36× lower in collision rate than ABS in diverse and challenging environments.

Briefly, we identify our contributions as follows:
\begin{enumerate}
    \item We propose an adaptive safety framework, BAS (Bridging Adaptivity and Safety), for legged locomotion.
    \item We propose an on-policy fine-tuning method to enhance the robustness of the parameter estimator in dynamic environments.
    \item We validate the adaptivity, safety, and agility of BAS through extensive evaluations in both simulation and real-world scenarios.
    \item We provide theoretical insights for parameter-conditioned reach-avoid value functions, which support the practical algorithms.
\end{enumerate}
    
\vspace{-10pt}
\section{Preliminaries and Problem Formulation}
\paragraph{Dynamics}
The dynamics is defined by state $s\in \mathcal{S}\subset \mathbb{R}^{|s|}$ and action $a\in \mathcal{A}\subset \mathbb{R}^{|a|}$ and environmental physical parameters as $e\in \mathcal{E}\subset \mathbb{R}^{|e|}$:
$s_{t+1} = s_t+f(s_t,a_t,e)$.
For simplicity\footnote{In analysis, we assume the environment is stationary, and $e$ is unknown but static. In experiments, $e$ can be time-variant.}, in this paper, we denote $e$ as the physical parameters, i.e., the combination of the mass of payload, the friction coefficient, the CoM shift, etc., which is assumed static within a trajectory in training sessions.
The observations are from proprioceptive and exteroceptive sensors, denoted as $o=h(s)$ where $h$ acts as the sensor mapping.
\paragraph{Goal Settings}
Given local position and goals $\mathcal{T}\in\Gamma$, we learn a goal-conditioned reaching policy $\pi:\mathcal{O}\times\Gamma\rightarrow\mathcal{A}$ to maximize the expected return: $J(\pi) = \mathbb{E}_{\pi,\mathcal{T}}\left[\sum_{t=0}^{\infty}\gamma_{RL}^t r(s_t,a_t,\mathcal{T})\right]
$, where $r(\cdot)$ is the reward at time $t$ and $\gamma_{RL}$ is the discount factor.
\paragraph{Safety Settings}
First, we denote the system trajectory starting from state \( s \) while using control inputs from the policy \( \pi \) under environmental parameter $e$ as \( \xi_{s}^{\pi,e}(\cdot): \mathbb{R} \rightarrow \mathcal{S} \). As in~\cite{bansal2017hamiltonjacobireachabilitybriefoverview}, we define several basic sets:
The target set \( \mathcal{T} \in \mathcal{S} \) which represents the area of the goal,
the constraint set \( \mathcal{K} \in \mathcal{S} \) which refers to the traversable areas for robots.
And the failure set \( \mathcal{F} = \mathcal{K}^C \), which is the complement of the constraint set and represents hazardous areas like obstacles.

Based on those basic sets above, we can define the following sets in the context of the reachability theory.
The safe set is defined as the set of states from which the robot can start and has a positive probability of rolling out a trajectory without failure, expressed as:
\(
\omega^{\pi,e}(\mathcal{F}) := \{s \in \mathcal{S} \mid \forall \tau \geq 0, \xi_{s}^{\pi,e}(\tau) \notin \mathcal{F}\}
\).
The backward reachable set is the collection of states from which the robot has a positive probability of reaching the target:
\(
\mathcal{R}^{\pi,e}(\mathcal{T}) := \{s \in \mathcal{S} \mid \exists \tau \geq 0, \xi_{s}^{\pi,e}(\tau) \in \mathcal{T}\}
\).
And the reach-avoid set combines the safe set and the backward reachable set:
\(
\mathcal{RA}^{\pi,e}(\mathcal{T}, \mathcal{F}) := \{s \in \mathcal{S} \mid \exists \tau \geq 0, \xi_{s}^{\pi,e}(\tau) \in \mathcal{T} \wedge \forall \tau \geq 0, \xi_{s}^{\pi,e}(\tau) \notin \mathcal{F}\}
\), which represents states from which the robot can reach the target while avoiding failure.

\paragraph{Reach-Avoid Value and Time-Discounted Reach-Avoid Bellman Equation (DRABE)}
Identical to the vanilla reach-avoid analysis~\citep{bansal2017hamiltonjacobireachabilitybriefoverview}, we define two Lipschitz-continuous functions $l(\cdot),\zeta(\cdot):\mathcal{S}\rightarrow\mathcal{R}$ which satisfy $\begin{cases}l(s)\leq 0 \iff s\in \mathcal{T}\\\zeta(s)>0\iff s\in \mathcal{F}\end{cases}$to illustrate if the robot has reached the target or collides with obstacles. 
Note that this function is only dependent on the state $s$ and is environment-agnostic. Then we define the reach-avoid value function $V^\pi_{RA}$ which satisfies $V^\pi_{RA}(s,e)\leq 0 \iff s\in \mathcal{RA}^{\pi,e}(\mathcal{T},\mathcal{F})$:
\begin{align}
    V^\pi_{RA}(s,e) = 
    \min_{\tau\in\{0,1,\dots\}}
    \max{\{l(\xi_{s}^{\pi,e}(\tau)), \max_{\kappa\in\{0,1,\dots,\tau\}}\zeta(\xi_{s}^{\pi,e}(\kappa))\}}~.
    \label{eq:ra_vf}
\end{align}
Note that a negative reach-avoid value guarantees a successful trajectory without collision till now, and a positive possibility to reach the target in the future. 
However, the function above is not learnable because it's not a contraction mapping, which fails to guarantee convergence in value iteration.
To learn this function, as introduced in \cite{Hsu2021}, we use Discounted Reach-Avoid Bellman Equation (DRABE) to make the value iteration a contraction mapping:
\begin{align}
    B_\gamma[V_{{RA}_\gamma}^\pi](s_t,e) = (1-\gamma)\max{\{l(s_t),\zeta(s_t)\}}
    +\gamma\max\{\min\{
    V^\pi_{RA_\gamma}({s_{t+1},e}),l(s_t)\},\zeta(s_t)\}
    \label{eq:ra_vf_td}
\end{align}
\cite{Hsu2021} also gives a mathematical proof of the DRABE operator $B_\gamma[\cdot]$ is a contraction mapping. And trivially, having $V^\pi_\gamma$ conditioned on a static physical parameter $e$ does not alter the proof, maintaining the guarantee of convergence.
\paragraph{Lipschitz-continuity of $V^\pi_\gamma$}
We deduct comprehensive analysis on $V^\pi_\gamma$'s convergence and Lipschitz-continuity as presented in \Cref{sec:app}, from which we imply that to guarantee Lipschitz-continuity of $V^\pi_\gamma$, $\pi$ should be not too sensitive to $s$ and $e$. To this end, we employ an L2-regularization and weight clipping on $\pi$ to lower its sensitivity to $s$ and $e$. 
Moreover, as~\cite{liu2021regularizationmatterspolicyoptimization} notes, regularization also matters in policy optimization in the context of RL because it provides better sampling complexity and return distribution.

\section{METHODOLOGIES}
\begin{figure*}[!t]
    \centering
    \includegraphics[width=\linewidth]{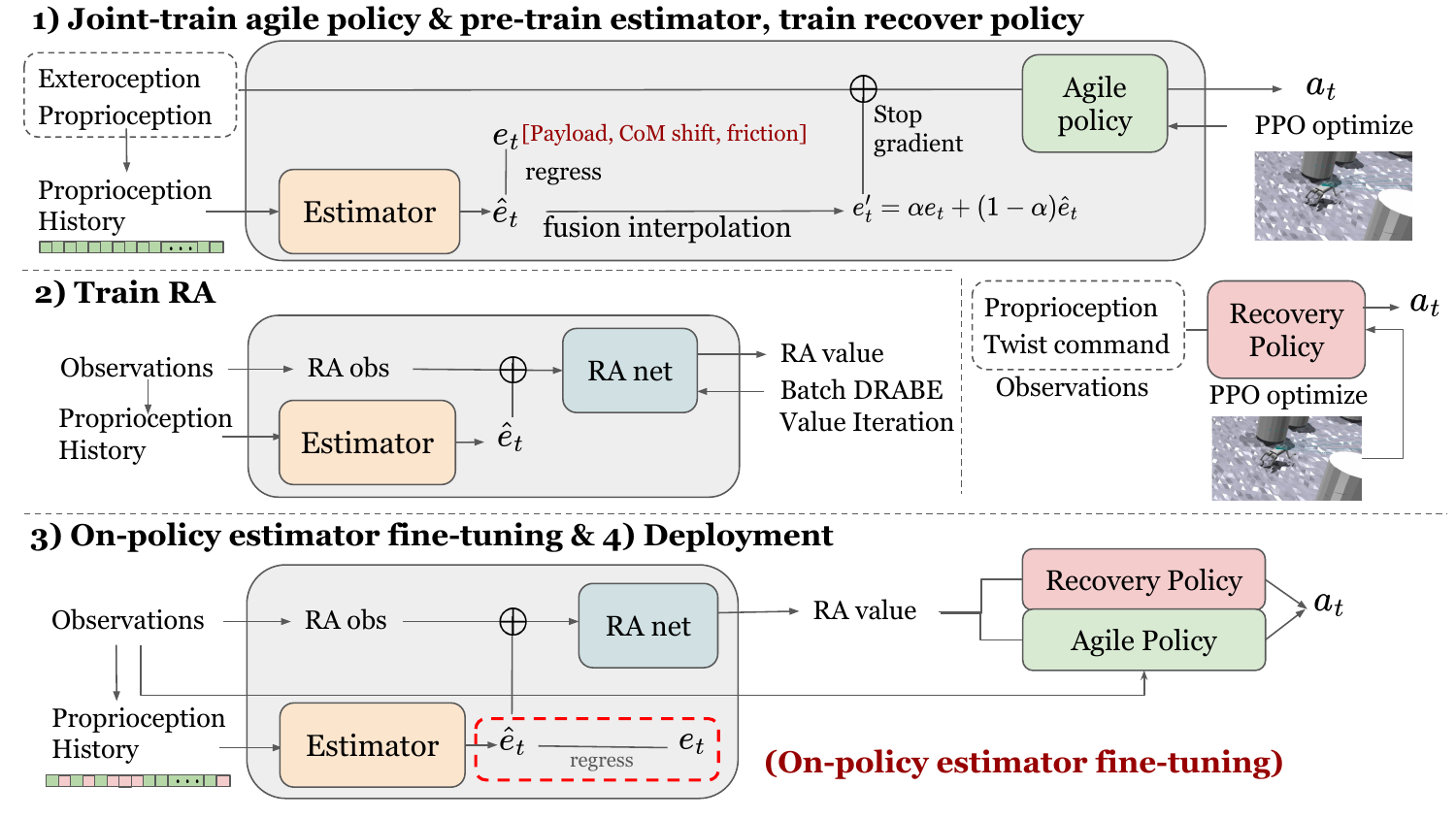}
    \caption{BAS Pipeline Overview. }
    \label{fig:ovr}
    \vspace{-10pt}
\end{figure*}

In this section, we present our proposed framework as shown in \Cref{fig:ovr}, which has four training phases:(\Cref{sec:phase_1}) training parameter estimator for adaptation; (\Cref{sec:phase_2}) training RA network;
(\Cref{sec:phase_3}) on-policy fine-tuning estimator to address the history distribution shift, and real-world deployment. Here we denote ground-truth physical parameters as $e_t$, the estimated ones as $\hat{e_t}$, and with fusion interpolation, we feed the policy with $e'_t=\alpha e_t+(1-\alpha)\hat{e_t}$ in training, where $\alpha=\min(2*\text{training rate},1)$ is the clipped training rate. For simplicity, in the following explanations, agile and recovery policies are accordingly denoted as $\pi_{agile}$ and $\pi_{recovery}$.

\vspace{-10pt}
\subsection{Phase 1: Joint-Train Agile Policy and Physical Parameter Estimator}
\label{sec:phase_1}
\paragraph{Policy-Conditioned Physical Parameter Estimator}
Since the environmental factors are often inaccessible in the real world, we tackle this challenge by learning a concurrent estimator $\phi^{\pi_{agile}}(o_{t:t-49})$ conditioned on $\pi_{agile}$ from robot proprioception history. which explicitly estimates the mass of the payload $m$, the position shift of CoM $\Delta x_c,\Delta y_c,\Delta z_c,$ and the friction coefficient $\mu$, which are critical for the daily use of autonomous robots.
However, training a general state estimator with high accuracy is super challenging, so we first opt to train a policy-conditioned estimator to lower the challenges and then propose fusion interpolation in the joint-train pipeline to boost the accuracy further.

Additionally, note that physical parameters are policy-invariant variables. So, compared to predicting dynamics that tangle with policies, predicting physical parameters is more suitable for cases where multiple policies are used together like \cite{he2024agilesafelearningcollisionfree,hoeller2023anymalparkourlearningagile}. What is more, estimating policy-invariant physical parameters partly lessens the potential issue of the history pattern misalignment when changing policies. To further reduce this effect, we also perform policy fine-tuning as described in \Cref{sec:phase_3} to maintain high accuracy at inference.

\paragraph{Policy Training}
\label{sec:fusion}
Following ABS, we maintain the two policy switching structures: $\pi_{agile}$ and $\pi_{recovery}$. $\pi_{agile}$ is a goal-reaching policy and takes control most of the time,
and we also retain $\pi_{recovery}$ designs from ABS, which tracks a given twist command.
To make $\pi_{agile}$ adaptive and aware of physics, we add the estimated physical parameters to its observation spaces as $a_t=\pi_{agile}(o_t,\hat{e_t})$, and as for $\pi_{recovery}$, we train a robust tracking policy that works in all the environments with strong domain randomizations~\Cref{tab:hyp}.
\paragraph{Training Pipeline}

To ensure that policy and estimator work well together, we propose a fusion interpolation on parameters such as $e'_t=\alpha e_t+(1-\alpha)\hat{e_t}$ in \Cref{fig:ovr} in the joint-train pipeline inspired by \cite{Ji2022}, where $\alpha$ is the training rate.
Within this fusion mechanism, the agile policy receives $e'_t=\alpha e_t+(1-\alpha)\hat{e_t}$ as input rather than $e_t$ nor $\hat{e_t}$, because we expect the agile policy to converge fast with the aid of ground truth privileged observations for the beginning steps, and we want the agile policy and estimator to co-adapt to each other's distribution when the policy converges. Moreover, such fusion mechanism reduces noise in $e_t$ introduced by imperfect estimation in training time, helping to train a more stable $\pi_{agile}$.
Furthermore, to lessen overfitting, we employ an MSE loss with L2 regularization on the estimator as well.
We also perform an ablation analysis in \Cref{sec:abl_esti} to validate this joint-training pipeline and fusion interpolation.
\vspace{-10pt}
\subsection{Phase 2: Learning Adaptive Reach-Avoid Network}
\label{sec:phase_2}
As in ABS, the RA network learns a reach-avoid value function as described in \Cref{eq:ra_vf_td}. To boost the RA network's adaptivity, we also extend the RA observation space with the estimated physical parameters to make it aware of physics. To simplify the training, we opt to learn a policy-conditioned, normalized, and adaptive RA value function $V^\pi_\gamma(s,\hat{e})$ as a safety guard. The guarding is triggered when $V^\pi_\gamma(s,\hat{e})>0$, and then the system calls $\pi_{recovery}$ to take control. 
Typically, the RA network is learned through the MSE loss to the target value from DRABE as $L=\frac{1}{T}\|\hat{V^\pi_\gamma}(s,\hat{e})-B_\gamma[V^\pi_\gamma(s,\hat{e})]\|^2 $ using $B_\gamma[\cdot]$ from \Cref{eq:ra_vf_td}.

\vspace{-10pt}
\subsection{Phase 3: On-Policy Estimator Fine-Tuning}
\label{sec:phase_3}
However, as our experiment shows in \ref{sec:abl_esti}, the estimation isn't accurate enough (e.g., $>$ 0.5kg error in mass). The cause probably lies in our structure which has two policies contribute to the same history buffer, potentially leading to a distribution shift on history, thus degrading the accuracy of the estimation.
To cope with the distribution shift to ensure that the estimator performs well during policy switching, we fine-tune it end-to-end with supervision in a deployment where $\pi_{agile}$, $\pi_{recovery}$ and RA network work together. 
Note that this is different from the training session in~\Cref{sec:phase_1} in that we generate the history rollouts solely with $\pi_{agile}$ in phase 1, but with both policies taking effect in turn at this phase. 

\section{EXPERIMENTS}

In this section, we present a series of simulation experiments in IsaacGym~\citep{rudin2022learningwalkminutesusing,makoviychuk2021isaacgymhighperformance} following the simulation setup and reward settings in ABS~\citep{he2024agilesafelearningcollisionfree} with domain randomization settings in \Cref{tab:hyp}. and real-world experiments on Unitree Go1 with onboard computations to investigate the following questions.

\textbf{Q1}: What are the most effective methodologies for achieving a balance between adaptive safety and agility in robotic systems?

\textbf{Q2}: What is the recipe for training the best estimation module in BAS?

\textbf{Q3}: How can we quantitatively assess the adaptivity and robustness of the BAS framework through in-depth analytical and experimental evaluations?

\textbf{Q4}: How well does BAS perform in real-world unseen scenarios, and how accurate can BAS's parameter estimator be in the real world?

\begin{table*}[h]
    \centering
    \scriptsize
    \begin{tabular}{llllll}

    \hline
    Policy &  Collision Rate(\%) $\downarrow$& Reach Rate(\%) $\uparrow$ & Timeout Rate(\%) & $\bar{v}_{peak}$ of success (m/s) $\uparrow$\\
    \hline
    \textbf{a) Adaptivity-wise}\\
    \hline
    BAS &   \textbf{1.11}&\textbf{93.84}&5.06&\textbf{2.70}\\
    BAS w/o explicit estimator  &  5.64 & 90.50 & 3.86 & 2.65\\
    ABS & 14.84 & 63.83 & 21.33 & 2.65\\
    RMA-RA & 12.51 & 80.12 & 7.37 & \textbf{2.70}\\
    Action-Distillation  &15.72& 68.99 & 15.29 & 2.63\\
    
    \hline
    \textbf{b) Safety-wise}\\
    \hline
        BAS &  \textbf{1.11}&\textbf{93.84}&5.06&2.70\\
        BAS-Lagrangian   &3.20& 90.40 & 6.40 & 2.51\\
        RMA-Lagrangian& 13.69&	76.33&	9.98&	2.48\\
        BAS-$\pi_{agile}$  &10.35& 89.00 & 0.65 & 2.75 \\
    \hline
    \textbf{c) For adaptivity-robustness analysis}\\
    \hline
    BAS &   1.11&93.84&5.06&2.70\\
    BAS-random &  100.00 & 0.00 & 0.00 & / \\
    RMA-RA &  12.51 & 80.12 & 7.37 & 2.70\\
    RMA-RA-random & 19.37 & 74.59 & 6.04 & 2.49\\
    \hline
    \end{tabular}
    \caption{Simulation experimental results. Collisionrefers to trajectories with collsion, Reach stands for reaching the target, and Timeout stands for being safe all over the trajectory without reaching the target. $\bar{v}_{peak}$ of success refers to the peak speed in a trajectory on average of all the successful ones.
    Note that ABS values may differ from~\cite{he2024agilesafelearningcollisionfree} because these experiments are done under larger domain randomizations, as shown in~\Cref{tab:hyp}.}
    \vspace{-15pt}
    \label{tab:sim-fulldr}
\end{table*}
\vspace{-5pt}    
\subsection{Safety and Agility Performance Analysis}
\label{sec:results}
\par To answer Q1 (\textit{What are the most effective methodologies to achieve a balance between adaptive safety and agility?}), we compare the non-collision rates and average top speed within a trajectory in the simulation between BAS and other adaptive and/or safe locomotion baselines. 
To show BAS's adaptivity, we introduce the following baselines:
\textbf{1) ABS}, which has non-adaptive $\pi_{agile}$ and RA network;
\textbf{2) BAS w/o explicit estimator}, which adopts long-short term history structures and learns an encoder that maps history to latent space with end-to-end RL training~\citep{li2024reinforcementlearningversatiledynamic};
\textbf{3) RMA-RA}, which incorporates RMA~\citep{kumar2021rmarapidmotoradaptation} and RA network with the latent environmental representation $z_t$ as the additional inputs.
\textbf{4) Action-Distillation}, which is similar to RMA and is inspired from~\cite{Lee2020}, where a student policy is distilled from an adaptive teacher policy by minimizing the difference between their actions.
\textbf{5) BAS-$\pi_{agile}$}, which only uses $\pi_{agile}$;
\textbf{6) BAS-Lagrangian}, which learns $\pi_{agile}$ with PPO-Lagrangian~\citep{lagrangianpolicy} with explicit estimation without RA network;
\textbf{7) RMA-Lagrangian}, which learns a teacher PPO-Lagrangian policy with RMA and then distills it into a student policy.

As shown in \Cref{tab:sim-fulldr} (a), BAS outperforms original ABS by 50\% in reach rates in varied physics and distinctively stands out with the lowest collision rate and the highest reach rate throughout all of the adaptive methods.
And in \Cref{tab:sim-fulldr} (b), the RA safeguard structure also outperforms policies trained by PPO-Lagrangian, especially in agility.
Moreover, validation of the effect of safeguarding on the entire system is observed through the comparison between BAS and BAS-$\pi_{agile}$, which indicates that adopting RA guard would transfer most of the failure cases to success cases or safe cases. 
\vspace{-12pt}
\subsection{Ablation Studies on Estimator}
\begin{figure}[h]
    \centering
    \includegraphics[width=\linewidth]{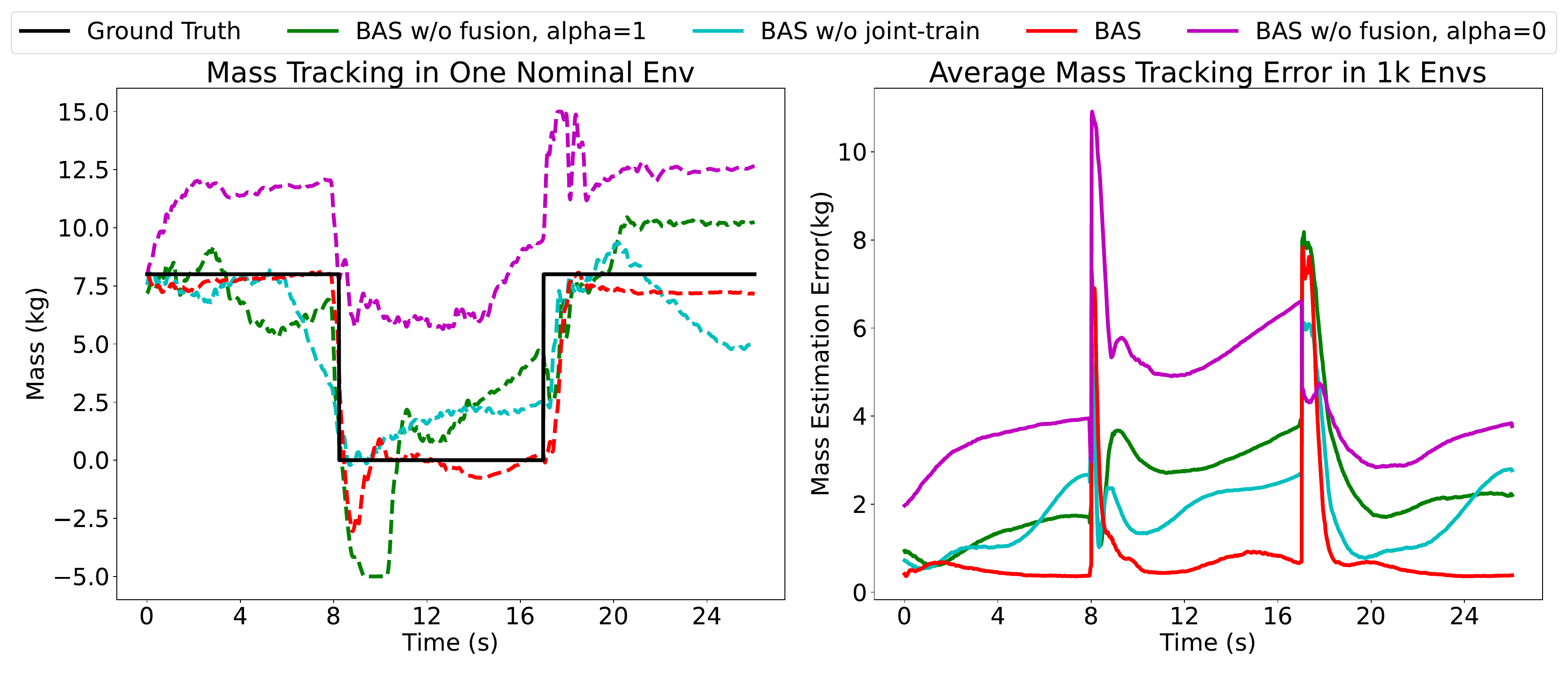}
    \vspace{-10pt}
    \caption{Mass estimation tracking of BAS, BAS w/o fusion and BAS w/o joint-train pipeline. Environment and the history buffer resets per 8s.}
    \label{fig:mass_est}
    \vspace{-10pt}
\end{figure}

To answer Q2 \textit{(How to train the best estimation module in BAS?)}, we investigate our two proposed methodologies to train the estimator: joint-train pipeline with fusion interpolation and on-policy fine-tuning.
\paragraph{Estimator Training Pipeline}
\label{sec:abl_esti}
\vspace{-6pt}
For ablation purposes, we test BAS\textbf{ w/o fusion} (arbitrarily setting $\alpha$ in \Cref{fig:ovr} to a constant 0 or 1) and BAS \textbf{w/o joint-train} (first learn a privileged policy then learn estimation from rollout data and use estimation as privileged observation at inference) as in \Cref{tab:saft}(a), which shows that the fusion interpolation offers a better accuracy on estimation and better overall agility-safety performance.
Also, we demonstrated the tracking of mass of payload as in \Cref{fig:mass_est}, where the joint-trained estimator is much more accurate. 
\vspace{-5pt}
\paragraph{On-Policy Fine-Tuning}
\label{sec:abl}
\begin{table*}[!h]
    \centering
    \scriptsize
    \begin{tabular}{lllllll}
    \hline
    Entry & estimation loss &  Collision Rate(\%) $\downarrow$& Reach Rate(\%) $\uparrow$ & Timeout Rate(\%) & $\bar{v}_{peak}$ of success (m/s)$\uparrow$\\
    \hline    
    \multicolumn{3}{l}{a) \textbf{Ablation: on training pipelines (before finetuning)}}\\
    \hline
    BAS & \textbf{0.570}& \textbf{3.10} & \textbf{92.48} & 4.42 & \textbf{2.69} \\
        BAS w/o fusion($\alpha\equiv1$) & 1.955& 3.71 & 91.10 & 5.19 & 2.66 \\
        BAS w/o fusion($\alpha\equiv0$) & 5.008 & 16.31 &52.30 & 31.39 & 2.63\\
        BAS w/o joint-train & 1.511 &6.21&88.20&1.89&\textbf{2.69}\\
    \hline
    \multicolumn{2}{l}{b) \textbf{Ablation: on-policy finetuning}}\\
    \hline
    BAS w/o finetuning &0.570& 3.10 & 92.48 & 4.42 & \textbf{2.69}\\
    BAS  &   \textbf{0.323} &\textbf{1.11}&\textbf{93.84}&5.06&2.68\\
    \hline
    \end{tabular}
    \caption{Comparisons on estimators w/ and w/o fusion or joint-train and on-policy finetuning.}
    \label{tab:saft}
    \vspace{-10pt}
\end{table*}

Note that the estimator is only trained with the rollout of $\pi_{agile}$, which may not have seen the trajectories contributed by the agile and recovery policies. To this end, we implemented on-policy post-finetuning on the estimator to diminish this distribution shift in an end-to-end scheme.
As can be seen in \Cref{tab:saft},
BAS outperforms BAS w/o finetuning in both estimation accuracy and safety performance.
    \vspace{-5pt}
\subsection{Adaptivity-Robustness Analysis}
\label{sec:abl_adapt}

\begin{figure*}[!b]
\centering
    \vspace{-10pt}
    \includegraphics[width=1.01\linewidth]{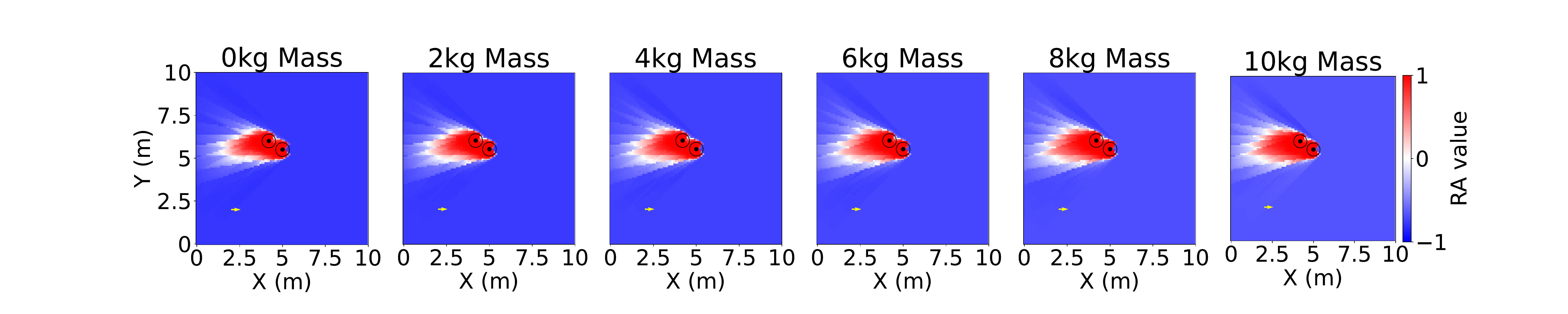}
    \vspace{-10pt}
    \caption{Heatmaps of RA values under the different mass of payloads at the state of 3.0m/s base linear velocity right forward. The more reddish, the higher the RA values, the more dangerous; the more bluish, the lower the RA values, the safer.}
    \label{fig:htmap}
    \vspace{-10pt}
\end{figure*}


\begin{wraptable}{r}{7cm}
\footnotesize
    \centering
    \vspace{-10pt}
    \begin{tabular}{c c}
    \hline
         Hyperparameter Name & Value \\
         \hline
         Mass of Payload range(kg) & -2.0,12.0\\
         Friction range& 0.25,1.5\\
         CoM shift-x(m) range& -0.05,0.05\\
         CoM shift-y(m) range& -0.05,0.05\\
         CoM shift-z(m) range& -0.05,0.15\\
         External Force-x range& -15N,15N\\
         External Force-y range& -15N,15N\\
         \hline
    \end{tabular}
    \vspace{-2mm}
    \caption{Domain Randomization Setting}
    \label{tab:hyp}
    \vspace{-0.4cm}
\end{wraptable}
To Answer Q3 (\textit{Can we identify adaptivity of BAS
with deeper analysis?}), we visualize the heatmap of RA values under different physical conditions (see~\Cref{fig:htmap}). The trend in RA values aligns with common sense: heavier payloads correlate with greater danger.

Moreover, for further adaptivity analysis, we try to compare BAS to the classic adaptive baseline, RMA. As conservativeness can be identified by robustness to noise in adaptation modules, we test \textbf{BAS-random} and \textbf{RMA-RA-random}, where the output of the adaptation module (explicit estimation $e_t$ in BAS, latent $z_t$ in RMA) is replaced with random numbers in the same distributions. 
As \Cref{tab:sim-fulldr} (c) shows, BAS deviates when the predicted mass is masked with random numbers, while masking RMA's latent vector has a minor loss in performance, which means that BAS is less conservative and more adaptive than RMA.
Explicit estimation also enhances the interpretability of the system by providing a clear understanding of the underlying physical significance of the estimations. Conversely, if environments are encoded to latent space, their meaning may remain obscure.

In summary, BAS outperforms other baselines in agility and safety metrics across our testing environments, demonstrating that its adaptivity, safety, and agility are all linked together.
\subsection{Real-World Experiments }
\begin{figure}
    \centering    
    \includegraphics[width=\linewidth]{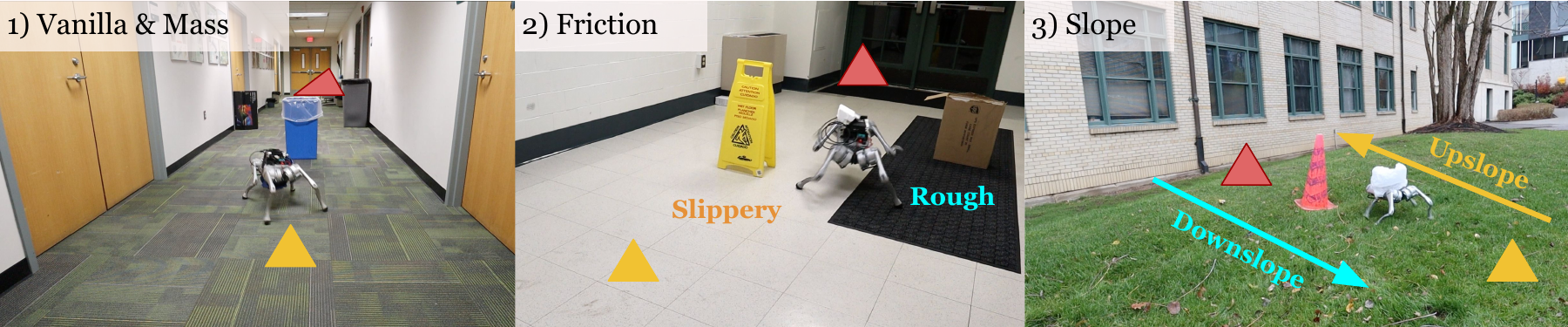}
    \caption{
    Real Experiment for Adaptive Safety test Settings, where \textcolor{orange}{yellow triangle} notes the starting point and \textcolor{red}{red triangle} notes the goal. Once the robot reaches the goal, we switch the goal and starting point. A trajectory from the start to the goal and then getting back to the start without collision is counted as success.
    \textbf{0) Vanilla test:} same as mass test settings, but without payloads.
    \textbf{1) Mass test}: carry a 5kg payload in a corridor and avoid boxes. \textbf{2) Friction test}: avoid box and a slip sign on very slippery floor and a dry carpet. \textbf{3) Slope test}: avoid a cone on a grass slope after rain, which is also very slippery.}
    \label{fig:setting_real}
\end{figure}

\paragraph{Experiment Setup}
To answer Q4 \textit{(How well does BAS perform in unseen scenarios in the real world, and how accurate is the parameter estimator of BAS in the real world?)},
we deploy our modules on a Unitree Go1 with onboard computations on NVIDIA Orin NX.
We test three entries here: BAS, ABS and RMA+Lagrangian, as described in \Cref{sec:results}, among which ABS is our prior work and RMA+Lagrangian is also an adaptive-and-safe baseline which is worth comparing.
In our experiment, the agility tests measure the agility of a policy under conditions as in \Cref{fig:setting_real} but without obstacles, and the safety tests quantify the safety by statistics on success rates in different environments. 
As shown in \Cref{fig:setting_real}, we have different environment settings tailored for each physical factor that should be adapted, together with the vanilla test. Note that CoM shift is very hard to identify in real world, so as an alternative, we build an overall test which is the slope test on grass to cover it.
\paragraph{Real World Safety-Agility Performances}
As shown in \Cref{tab:real}, BAS outperforms ABS and RMA+lagrangian in both safety and agility across different physics and settings. We also find that ABS fails to turn 90 degrees with a 5kg payload and struggles to maintain safety on slippery terrains during experiments. During experiments, we also encountered some failures with BAS. And we analyzed some causes of failure:
1) Restricted by limited visual angle. 2) The robot only encounters with mild collision with obstacles. 3) Indoor environments are obstacle-dense, and the ray-prediction network deviates due to out-of-distribution rays.
\begin{table}[h]
    \centering
    \small
    \begin{tabular}{|l|cccc|c|cccc|c|}
    \hline
       \multirow{2}{*}{Policy} & \multicolumn{5}{|c|}{\textbf{Adaptive Agility test(s)$\downarrow$}} & \multicolumn{5}{|c|}{{\textbf{Adaptive Safety test$\uparrow$}}}\\
        \cline{2-11}
        & Vanilla & Mass & Slope & Friction&Avg. & Vanilla & Mass &Slope & Friction&Avg.\\
        \hline
         BAS& \textbf{1.39}&  \textbf{1.67}& \textbf{1.50}& \textbf{1.09}&\textbf{1.41}& \textbf{ 8/8}&  \textbf{7/8}& \textbf{5/8} & \textbf{6/8}&\textbf{81.25\%}\\
        ABS & 1.52&  2.37& 1.67 & 1.40&1.74&  
        7/8&  1/8& 3/8 & 0/8&34.38\%\\
       
        RMA-Lag& 1.76&1.92  & 1.85& 2.02& 1.89& 6/8&  5/8& 0/8 & 2/8&40.63\%\\
    \hline
    \end{tabular}
    \caption{Test results in real world. For pure agility tests, we compare the average time consumed to run 2.4m from stance in 3 trials. For safety-related tests, we compare the average success rate of 8 trials. }
    \label{tab:real}
\end{table}
\paragraph{Real World Adaptation Analysis and Run-time Estimation}
\begin{figure}[!h]
    \centering
    \includegraphics[width=\linewidth]{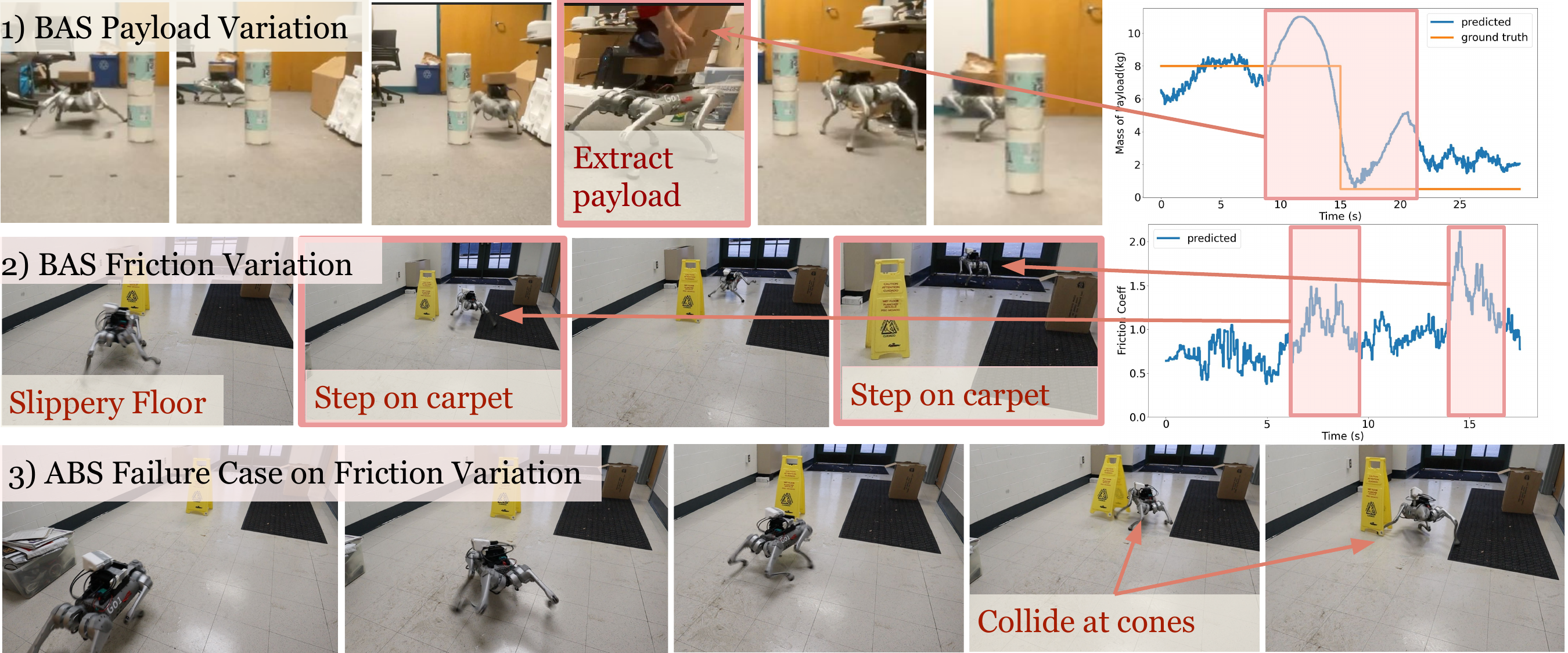}
    \caption{Adaptation analysis in real with online changes in the environment. 1) BAS Accomplishing collision avoidance while carrying an 8kg payload at first and then no payload in one trajectory. 2) BAS accomplishes collision avoidance in terrains with different frictions(liquid soap and water on the floor and dry mattress), while 3) ABS fails due to the lack of adaptivity.
    BAS estimator functions well in both cases with a correct trend.}
    \label{fig:real_adapt}
    \vspace{-4mm}
\end{figure}
\Cref{fig:real_adapt} shows that BAS maintains adaptive safety even under sudden environmental changes online, such as extracting the 8kg payload or sudden changes in terrain properties such as friction, while ABS fails with insufficient adaptivity to maintain safety in this case.
Moreover, as shown in the estimation plots in \Cref{fig:real_adapt}, the estimation remains accurate after the changes, and BAS accomplishes avoiding obstacles under all environmental conditions, confirming its adaptability. Note that the estimated values may differ from real-world ground truth because of imperfect simulation, especially in friction, so we mainly focus on the relative values and the trends here.
\vspace{-10pt}



\vspace{-3pt}
\section{CONCLUSIONS, LIMITATIONS, AND FUTURE PROSPECTS}
\vspace{-6pt}
In this paper, we propose BAS, which achieves collision-free locomotion in real-world dynamic environments and strikes a balance between adaptivity, agility, and safety by learning a nominal physical parameter estimator. 
For future works, we have several interesting research topics based on BAS: 1) BAS currently uses the 2D ray distances to the obstacles as the exteroceptive observation, and we may try to tackle 3D scenarios with VLAs\citep{zhen20243dvla3dvisionlanguageactiongenerative,kim2024openvlaopensourcevisionlanguageactionmodel} in the future; 2) We only trained and tested with static obstacles in this work, and will try to avoid highly dynamic obstacles in the future; 3) The current framework is focused on local obstacle avoidance, and we will try to combine high-level planning with low-level safety adaptation to accomplish more complex navigation problems in dynamic and challenging scenarios.
\vspace{-10pt}

\acks{
We gratefully acknowledge the dedication and contribution of Guanqi He, who helped us
repair the hardware and gave bunches of advice on hardware designs. We appreciate Haotian Lin, Wenli Xiao, and Jiawei Gao for their assistance in real-world experiments. Special thanks to Andrea Bajcsy and Toru Lin for their ideas in graphics design.
}
\bibliography{ref}

\appendix
\section{Detailed Proof}
\label{sec:app}
Since~\cite{li2024infinitehorizonreachavoidzerosumgames} proves that $V_\gamma(s)$ is Lipschitz-continuous to $s$, we extend the proof to prove that $V_\gamma^\pi(s,e)$ is Lipschitz-continuous to both $s$ and $e$. For $s$, introducing a static $e$ trivially doesn't alter the Lipschitz continuity of $V^\pi_\gamma$ to $s$.
So in this section we try to prove the Lipschitz continuity of the value function $V_\gamma^\pi(s,e)$ to environment factor $e$ in \Cref{app:Lipschitz}. 
\begin{theorem}{\textbf{(Lipschitz Continuity of $V^\pi_\gamma$ to $e$)} The Learned Value Function \( V^\pi_\gamma(s,e) \) Possesses Lipschitz Continuity w.r.t. Environmental Factors \( e \)}
under the following conditions:
\label{app:Lipschitz}
\begin{itemize}
    \item The functions \( l(s) \) and \( \zeta(s) \) are defined as \( L_l \)- and \( L_\zeta \)-Lipschitz continuous functions of the state \( s \).
    \item 
    $\gamma(1+L_{f_\pi})<1$, which will be naturally introduced in the proof.
    \item Given a specific policy \( \pi \), the transition dynamics defined as \( f_\pi(s,e) \mathrel{\mathop:}=f(s,\pi(s,e),e) \) are \( L_{f_\pi} \)-Lipschitz continuous w.r.t. the tuple \( (s,e) \). \\
    $L_{f_\pi}$ is defined as, for any states \( s_1, s_2 \in\mathcal{S}\) and environmental factors \( e_1, e_2\in\mathcal{E} \)
    \[
    \|f(s_1,\pi(s_1,e_1),e_1) - f(s_2,\pi(s_2,e_2),e_2)\| \leq L_{f_\pi}(\|e_1-e_2\| + \|s_1-s_2\|),
    \]
    where \( L_{f_\pi} \) is conditioned upon the policy \( \pi \). 

\end{itemize}
\end{theorem}
\noindent\textbf{\textit{Remark.}} As noted by~\cite{gouk2020regularisationneuralnetworksenforcing}, the sample complexity of neural network approximation can be enhanced if the function being approximated is continuous. Consequently, the Lipschitz continuity of the value function~\cref{eq:ra_vf_td} is a valuable property that leads to reliable empirical performance when using neural network approximations for Reach-Avoid values.
which can be found in \Cref{sec:app}.
\begin{proof}
    Here we note $V$ as for $V^\pi_\gamma$ because there's only one value function in this section.
    By definition, we got
    \begin{align*}
        V(s,e):= \min_{\tau\in\{0,1,\dots\}}{\max{\{\gamma^\tau l(\xi_{s}^{\pi,e}(\tau),
        \max_{\kappa\in\{0,1,\dots,\tau\}}\gamma^\kappa\zeta(\xi_{s}^{\pi,e}(\kappa)))\}}}
    \end{align*}
    And define $P(s,e,t)$ as payoff at timestep $t$:
    \begin{align*}
        &P(s,e,t):=\max{\{\gamma^t l(\xi_{s}^{\pi,e}(t),\max_{\kappa\in\{0,1,\dots,t\}}\gamma^\kappa\zeta(\xi_{s}^{\pi,e}(\kappa)))\}}
    \end{align*}
    For all $e_1,e_2\in \mathcal{E}$ and $s\in \mathcal{S}$, and $\theta>0$ we have:
    \begin{equation}
   \begin{cases}
        \forall t\in\mathcal{R}, P(s,e_1,t) > V(s,e_1)-\theta\\
        \exists \bar{t}\in\mathcal{R}, P(s,e_2,\bar{t}) < V(s,e_2)+\theta
    \end{cases} 
    \end{equation}
    Combining the two inequalities:  
    \begin{align*}
        &V(s,e_1)-V(s,e_2)-2\theta< P(s,e_1,\bar{t})-P(s,e_2,\bar{t})\\
        &\leq \max\{\gamma^{\bar{t}} L_l\|\xi_{s}^{\pi,e_1}(\bar{t})-\xi_{s}^{\pi,e_2}(\bar{t})\|, \max_{\kappa\in\{0,1,\dots,\bar{t}\}}\gamma^\kappa L_\zeta\|\xi_{s}^{\pi,e_2}(\kappa)-\xi_{s}^{\pi,e_2}(\kappa)\|\}
    \end{align*}
    We use $\Delta\xi(t):=\xi_{s}^{\pi,e_1}(t)-\xi_{s}^{\pi,e_2}(t)$, and by definition of $f$ 's Lipschitz continuity, we have
    \begin{align*}
        \Delta\xi(\bar{t})
        &\leq (1+L_f)\|\Delta\xi(\bar{t}-1)\|+L_f(\|p_1-p_2\|)\\
        &\leq\dots= ((1+L_{f_\pi})^{\bar{t}}-1)\|e_1-e_2\|
    \end{align*}
    Because this holds for some $\bar{t}$, so it must be less than the maximum for all $\bar{t}$. Thus we got
    \begin{align}
        V(s,e_1)-V(s,e_2)&\leq 2\theta +
        \max\{L_l,L_\zeta\} \max_{\bar{t}}{\{\max_{t\in\{0,1,\dots,\bar{t}\}}\gamma^{t}((1+L_{f_\pi})^{t}-1)\}\|e_1-e_2\|}
        \label{eq:continuity}
    \end{align}
    As $\theta$ is an arbitrary variable, we can set it to infinitesimal.
    To guarantee a finite bound for Lipschitz continuity of $V$ to $e$, it should be assured that $\gamma(1+L_{f_\pi})\leq1$ which necessarily holds that the Lipschitz constant is finite.
    Then we got the Lipschitz constant for the Value Funtion $V$ to environmental factor $e$:
    $$
    \forall s \in \mathcal{S}, V(s,e_1)-V(s,e_2)\leq L_V\|e_1-e_2\|~,    
    $$where
    $$
    L_V=\max\{L_l,L_\zeta\}\max_{t=0,1,\dots,T}\gamma^{t}((1+L_{f_\pi})^{t}-1)~,
    $$
    where $T$ denotes the maximum time steps for a system trajectory. Assuming $T\rightarrow\infty$ for infinite-horizon cases,
    by calculating the maximum point of $t$ in the right part, we got the upper bound of $L_V$:
    \begin{align}
        UB(L_V)=\max\{L_\zeta,L_l\}\cdot L_{f_\pi}\gamma^{t^*}\frac{\log(1+L_{f_\pi})}{-\log(\gamma(1+L_{f_\pi}))}~,
    \end{align}
    where
    $$t^*:=\frac{\log(\frac{\log(\gamma)}{\log(\gamma(1+L_{f_\pi}))})}{\log(1+L_{f_\pi})}~.$$
    Similar to \Cref{eq:continuity}, we can show that $V(s,e_2)-V(s,e_1)\leq 2\theta + L_V\|e_1-e_2\|$. Combining these two inequalities together, it can be implied that $V(s,e)$ is $L_V$-Lipschitz-continuous to $e$.
\end{proof} 
Following the proof, we can observe that $L_V$ is also bounded by $\max\{L_l,L_\zeta\}$ because the exponential term $\gamma^{t}((1+L_{f_\pi})^{t}-1)$ should be less than 1. 

The assumption $\gamma(1+L_{f_\pi})<1$ also gives a constraint that the dynamics with respects to the policy $\pi$ shouldn't be too sensitive to environment factor $e$, i.e. $\pi$ should be a robust policy to $e$.


\end{document}